\documentclass{article}
\usepackage{spconf,amsmath,graphicx}
\usepackage{ctable} % for \specialrule command
\usepackage{amsfonts}       % blackboard math symbols
\usepackage{amsthm}
\usepackage{subcaption}
\usepackage{algorithm}
\usepackage{algpseudocode}
\usepackage{hyperref}
\usepackage{enumitem}
\setlist{nosep, leftmargin=14pt}

\usepackage{mwe} % to get dummy images
\title{Robust White Matter Hyperintensity Segmentation on Unseen Domain}
\name{\parbox{\linewidth}{\centering Xingchen Zhao$^1$  \quad  Anthony Sicilia$^2$ \quad Davneet S. Minhas$^3$ \quad Erin E. O’Connor$^5$\\ \textit{Howard J. Aizenstein}$^4$ \quad \textit{William E. Klunk}$^4$ \quad \textit{Dana L. Tudorascu}$^4$ \quad \textit{Seong Jae Hwang}$^{1,2}$}}
\address{
    $^{1}$Intelligent Systems Program - University of Pittsburgh \\
    Department of $^{2}$Computer Science, $^{3}$Radiology, $^{4}$Psychiatry - University of Pittsburgh \\
    $^{5}$Department of Diagnostic Radiology \& Nuclear Medicine -  University of Maryland, Baltimore}
\begin{document}
%\ninept
%
\maketitle
\begin{abstract}
Typical machine learning frameworks heavily rely on an underlying assumption that training and test data follow the same distribution. In medical imaging which increasingly begun acquiring datasets from multiple sites or scanners, this identical distribution assumption often fails to hold due to systematic variability induced by site or scanner dependent factors.
Therefore, we cannot simply expect a model trained on a given dataset to consistently work well, or generalize, on a dataset from another distribution. In this work, we address this problem, investigating the application of machine learning models to unseen medical imaging data. Specifically, we consider the challenging case of Domain Generalization (DG) where we train a model without any knowledge about the testing distribution. That is, we train on samples from a set of distributions (sources) and test on samples from a new, unseen distribution (target). We focus on the task of white matter hyperintensity (WMH) prediction using the multi-site WMH Segmentation Challenge dataset and our local in-house dataset. We identify how two mechanically distinct DG approaches, namely domain adversarial learning and mix-up, have theoretical synergy. Then, we show drastic improvements of WMH prediction on an unseen target domain.

\end{abstract}
\begin{keywords}
Domain Generalization, Image Segmentation, Deep Learning, White Matter Hyperintensity
\end{keywords}
%

% \vspace{-5pt}
\section{Introduction} %~0.6 page
\label{sec:intro}
% \vspace{-10pt}
In traditional machine learning, an underlying assumption is that the model is trained on training data that is well representative of the testing data. That is, both the training and testing samples come from an identical distribution. However, this assumption has become difficult to satisfy in the modern medical imaging analysis community as it rapidly grows with multiple sites or scanners \cite{kuijf2019standardized,glocker2019machine}. Notably, datasets often exhibit heterogeneity (e.g., differing distributions of intensity) due to systematic variability induced by various site/scanner dependent factors, and commonly, imaging protocols.
{\let\thefootnote\relax\footnotetext{\hspace{-10pt}Accepted to IEEE International Symposium on Biomedical Imaging 2021}}

In this work, we view sites or scanners as distinct distributions or \textit{domains} and consider learning in the presence of multiple domains where the identical distribution assumption no longer holds.
Under this new training regime, a model is trained on data arising from some sites/scanners and tested on samples from a \textit{new} site/scanner \textit{unseen} during the training process.
Formally, at train-time, we observe $k$ domains (sites/scanners) refer to as \textit{sources} which have distributions $\mathbb{P}_1, \mathbb{P}_2, \ldots, \mathbb{P}_k$ over some space $\mathcal{X}$.
At test-time, we test the model on a distinct \textit{target} domain with distribution $\mathbb{Q}$ over $\mathcal{X}$.

This multi-domain construct appears in the literature in several medical image segmentation problems.
When a model pretrained on sources is further trained on additional target samples with labels, this is often referred to as Transfer Learning \cite{shin2016deep}.
Adding constraints to this setup, \textit{Domain Adaptation} (DA) assumes access to samples from $\mathbb{Q}$ but \textit{without} their labels \cite{li2020domain,ganin2015unsupervised}. This is a prevalent situation for tasks requiring costly data annotations (e.g., manual tracing of brain lesions).
Without labels, DA techniques may still utilize the target information to learn domain-agnostic features \cite{li2020domain,scannell2020domainadversarial,panfilov2019improving}, so that task performance is promising irrespective of the input domain. Importantly, these solutions still assume unlabeled data from $\mathbb{Q}$, and this assumption of pre-existing knowledge may be too strong in some practical scenarios.
% or transforming target samples to be more ``source-like" \cite{hoffman2018cycada}.
%Although solving for DA adds a great deal of generalizability to the models for medical imaging data, their success still depends on the amount of preexisting knowledge about the target $\mathbb{Q}$.

We therefore consider the problem of \textit{Domain Generalization} (DG) where we make \textit{no} assumptions about the target distribution, i.e., no access to the samples from $\mathbb{Q}$ during training.
This recently developing construct has been deemed more challenging, but also, more useful to real world problems where we want to generalize without any data from our target \cite{matsuura2019domain,sicilia2021domain}.
Yet, while a few DG approaches exist \cite{zhang2020generalizing,khandelwal2020domain}, DG is still a nascent concept in medical imaging especially on segmentation problems.
Several reasons contribute to this hindered progress, including the lack of theoretical understanding of the existing DG methods and their non-trivial adaptation to segmentation models such as U-Net.
Nonetheless, a domain generalizable model is expected to bring practical benefits: retrospectively, we may better leverage existing siloed multi-domain datasets, and prospectively, we can reliably use these models on unseen datasets.

\textbf{Contribution.} In this work, we %propose a domain generalizable model for the task of white matter hyperintensity segmentation: 
%and 
ask the following question: \textit{Can we devise a segmentation model that generalizes well to unseen data?} We answer this as follows: 
\textbf{(1)} We investigate two mechanically different DG methods, namely domain adversarial neural network and mixup, and identify their theoretical commonality. \textbf{(2)} We use these DG methods to build upon a U-Net segmentation model, tackling the WMH segmentation problem on a multi-site WMH Challenge Dataset \cite{kuijf2019standardized} and drastically improving performance of a traditional U-Net on the DG task. \textbf{(3)} We further test our model on our local data which is not a part of the aforementioned multi-site dataset. We make our code publicly available.\footnote{ \textcolor{blue}{\scriptsize\url{https://github.com/xingchenzhao/MixDANN}}}

% \section{Related Work} % combine with intro
% \label{sec:related}
% \input{related.tex}

% \vspace{-3pt}
\section{Methods} %~ 1 page
\label{sec:methods}
% \vspace{-5pt}
Intuition tells us that if our features are invariant to the domain, then the main task should not be affected by the domain of the input. In fact, recent theoretical argument \cite{sicilia2021domain} formally suggests such domain invariance in the feature space as a solution for DG. This motivates our proposed approach.
We first employ the common domain adversarial training algorithm \textit{DANN} which learns domain invariant features that fool a domain discriminator.
We further show the data-augmentation algorithm \textit{mixup} \cite{zhang2018mixup} may also be viewed as promoting domain invariance. We then propose to use both DANN and mixup after identifying their theoretical connection in DG.

% while the original mixup paper also mixes across y_i, our segmentation setting is slightly different due to the use of a dsc score loss. We adopt a loss balancing strategy similar to \cite{that one paper} with implementation details in the code.

% \vspace{-5pt}
\subsection{Domain Adversarial Neural Network for DG}
% \vspace{-5pt}

Based on the seminal theoretical work of Ben-David \textit{et al.} \cite{ben2007analysis, ben2010theory}, Ganin and Lempitsky \cite{ganin2015unsupervised} proposed the commonly used algorithm DANN which learns domain invariant feature representations as desired. This algorithm breaks the model used for the task into two components: a feature extractor $r_\theta$ parameterized by $\theta$ and a task-specific network $c_\sigma$ parameterized by $\sigma$. In addition to these, we also train a domain discriminator $d_\mu$ to classify from which domain each data point is drawn. To learn a domain invariant representation, the feature extractor is trained to \textit{fool} this domain discriminator. Intuitively, if $d_\mu$ cannot identify the domain, then the feature representation learned by $r_\theta$ must be void of domain-specific features. In more detail, we may write the DANN objective computed for multiple source domains $\mathbb{P}_1, \ldots, \mathbb{P}_k$ as below
\begin{equation}\small
\label{eqn:dann_obj}
% \begin{split}
 \min_{\sigma, \theta} \max_\mu \mathbf{E}_{(x_i)_i \sim (\mathbb{P}_i)_i} \left [\sum\nolimits_{i}\mathcal{L}_T(\sigma, \theta, x_i) \right. - \gamma \left .\sum\nolimits_{i} \mathcal{L}_D(\mu, \theta, x_i) \right]
% \end{split}
\end{equation}
where $\mathcal{L}_D$ is the cross-entropy loss for domains
\begin{equation}\small
  -\mathcal{L}_D(\mu, \theta, x) = \sum\nolimits_j \mathbb{I}[x \sim \mathbb{P}_j] \log ((d_\mu \circ r_\theta (x))_j)   
\end{equation}
and $\mathcal{L}_T$ is a task specific loss (e.g., in our segmentation setting, this might be the DSC loss). From this objective, the learned model $c_\sigma \circ r_\theta$ may be adept at the task (i.e., by minimizing $\mathcal{L}_T$), but also invariant of domains (i.e., by maximizing $\mathcal{L}_D$). As is usual, we optimize this objective by simultaneous gradient descent implemented by inserting a Gradient Reversal Layer \cite{ganin2015unsupervised} between $d_\mu$ and $r_\theta$.

\begin{figure*}
    \centering
    \includegraphics[width=0.8\textwidth]{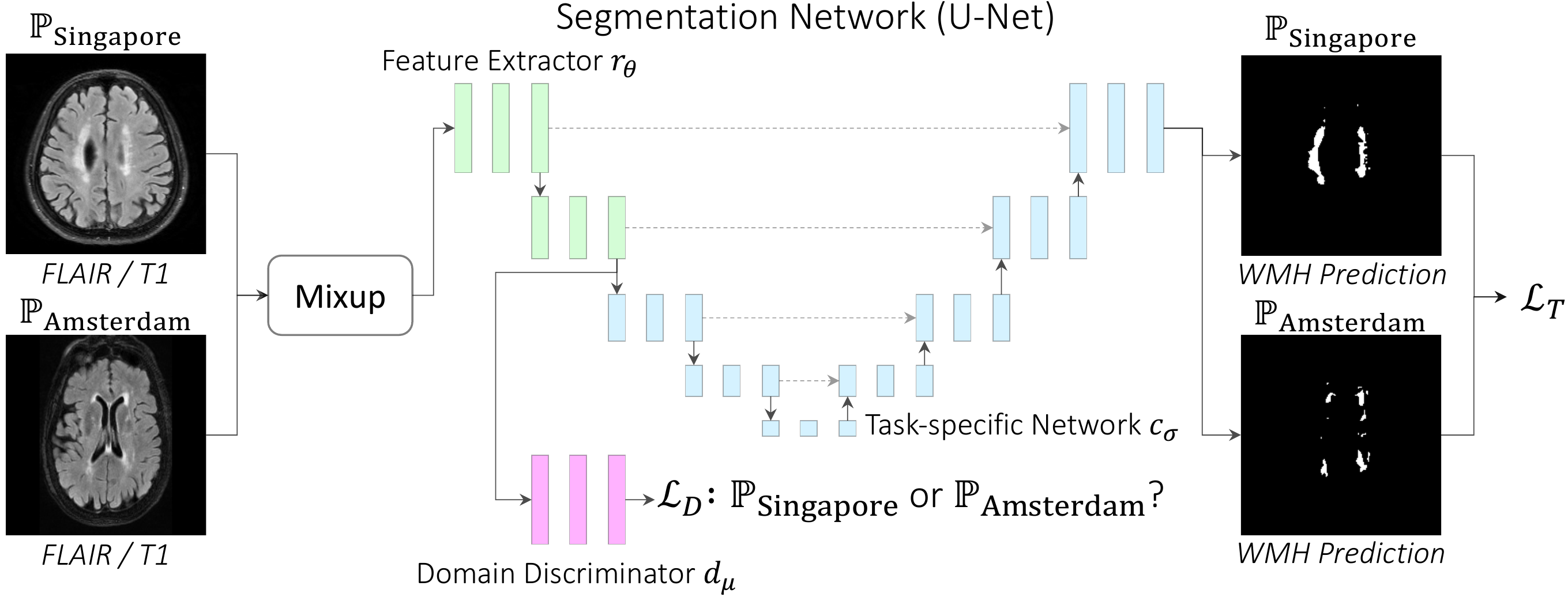}
    % \vspace{-10pt}
    \caption{MixDANN with Amsterdam and Singapore sources. }
    \label{fig:model}
    % \vspace{-15pt}
\end{figure*}

More specific to our segmentation task, it is unclear how to break up a fully convolutional neural network into a \textit{feature extraction} component and a \textit{task-specific} component.
% I need an author to cite here.
Motivated by \cite{shirokikh2020unet}, suggesting domain information is typically found in the earlier convolutional layers of a network, we generally limit $r_\theta$ to a few blocks in the \textit{downward} path of a U-Net (Fig.~\ref{fig:model}). We provide exact details in the code.
% \textcolor{red}{Perhaps more details here.}

% \vspace{-5pt}
\subsection{Mixup}
% \vspace{-5pt}
Besides application of DANN, we also propose to use the common data-augmentation algorithm Mixup \cite{zhang2018mixup}. At first glance, this extension is fairly simple. Still, in the presence of multiple domains $\mathbb{P}_1, \ldots, \mathbb{P}_k$, this algorithm complements the optimization objective in Eq.~\eqref{eqn:dann_obj} because it also aims to produce domain invariance in our learning algorithm (discussed in the next sub-section). The algorithm is defined as follows. Suppose we have a batch of data-points $\{x_i\}_i$ with $x_p$ and $x_q$ two distinct data-points from this sample. Further, let $\lambda \sim \mathrm{Beta}(\alpha, \alpha)$. Then, we define the new \textit{mixed} data-point: 
\begin{equation}\small
\label{eqn:mixup}
  \tilde{x}_{p,q} = \lambda  x_p + (1 - \lambda )  x_q. 
\end{equation}
With a certain probability (e.g., 0.5 in all our experiments), we can then substitute every data-point in the batch $\{x_i\}_i$ with a \textit{mixed} counterpart by randomly pairing the elements of $\{x_i\}_i$ and using Eq.~\eqref{eqn:mixup} to combine them. We do remark that the original proposal for mixup also mixes across the label space. Our segmentation setting is slightly different due to the common use of a Dice score loss. Thus, we adopt a loss balancing strategy similar to \cite{panfilov2019improving}: $\mathcal{L}(\tilde{x}_{p,q}) = \lambda\mathcal{L}^{p}(\tilde{x}_{p,q}) + (1-\lambda)\mathcal{L}^{q}(\tilde{x}_{p,q})$ where $\mathcal{L}^{i}$ is the loss with the label $y_i$ of $x_i$.

% \vspace{-5pt}
\subsection{Theoretical Motivations}
% \vspace{-5pt}

\noindent\textbf{DANN and the $\mathcal{H}$-divergence.}
While many works have used the motivation of domain invariant features for DG \cite{matsuura2019domain, li2018domain_mmd}, we note that the original theoretical motivation of DANN was based on the \textit{domain adaptation} theory proposed by Ben-David \textit{et al.} \cite{ben2010theory}. Recent extensions \cite{matsuura2019domain, sicilia2021domain, albuquerque2019adversarial} build upon existing theory and algorithm to demonstrate justified applications of DANN in \textit{DG}. Theoretically speaking, upper bounds on the error of the unseen target suggest we should minimize the $\mathcal{H}$-divergence -- a measure of the \textit{difference} between two domains. %-- appears in an upper bound of the error on an unseen target. 
% Therefore, to proxy minimization of the unseen target error, we can minimize the $\mathcal{H}$-divergence in the upper-bound. %(amongst other quantities, such as the error on the sources). 
The objective described in Eq.~\eqref{eqn:dann_obj} can be interpreted as minimizing the $\mathcal{H}$-divergence in a fairly formal sense because the $\mathcal{H}$-divergence measures a classifier's ability to distinguish between domains. Motivated by this, we learn invariant features in Eq.~\eqref{eqn:dann_obj} by maximizing the errors of the domain classifier $d_\mu$.

\vspace{5pt}
\noindent\textbf{A Formal Discussion of Mixup.} Mixup \cite{zhang2018mixup} is a special case of \textit{Vicinal Risk Minimization } \cite{chapelle2001vicinal}. Usually, in machine learning, we use \textit{Empirical Risk Minimization} which suggests estimating the true data-distribution by the empirical measure $P(x,y) = \frac{1}{n}\sum\nolimits_{i=1}^n \delta_{x_i}(x)\delta_{y_i}(y),$
% \begin{equation}\small
%     P(x,y) = \frac{1}{n}\sum\nolimits_{i=1}^n \delta_{x_i}(x)\delta_{y_i}(y),
% \end{equation}
with the Dirac measure $\delta_*$ estimating density at each data-point. In the more general case of Vicinal Risk Minimization, we allow freedom to use a density estimate in the \textit{vicinity} of the data-point $x_i$ using the vicinal measure $v_*$ as $P_v(x,y) = \frac{1}{n} \sum\nolimits_{i=1}^n v_{x_i}(x) \delta_{y_i}(y)$.
% \begin{equation}\small
%     P_v(x,y) = \frac{1}{n} \sum\nolimits_{i=1}^n v_{x_i}(x) \delta_{y_i}(y)
% \end{equation}
In Mixup, the modifications to our data described in Eq.~\eqref{eqn:mixup} equate to sampling from a certain vicinal distribution $u_*$: $u_{x_i}(x) = \frac{1}{n} \sum\nolimits_{j=1}^n  \mathbf{E}_\lambda \left [\delta_{\lambda x_i + (1-\lambda) x_j} (x)\right]$
% \begin{equation}\small
%     u_{x_i}(x) = \frac{1}{n} \sum\nolimits_{j=1}^n  \mathbf{E}_\lambda \left [\delta_{\lambda x_i + (1-\lambda) x_j} (x)\right]
% \end{equation}
where $\lambda \sim Beta$.
% follows a Beta distribution.

% \setlength{\textfloatsep}{5pt}% Remove \textfloatsep
\algrenewcommand\algorithmicindent{0.5em}%
\begin{algorithm}[t!] \footnotesize
    \caption{MixDANN for Domain Generalization}
    \label{alg:one}
    \textbf{Data}: Set of sample $x$ and label $y$ from Sources $\mathbb{P}_1,\dots,\mathbb{P}_k$ and Target $\mathbb{Q}$\\
    \textbf{Models}: U-Net (Feature Extractor $r_\theta$ and Task-specific Network $c_\sigma$), Domain Discriminator $d_\mu$
    % \textbf{Other Parameters:  }
    \begin{algorithmic}[1] % The number tells where the line numbering should start
        \Procedure{Train MixDANN using Sources $(\mathbb{P}_i)_i$}{}
            \For{mini-batch of samples from sources $(x_i)_i \sim (\mathbb{P}_i)_i$} 
                \State{\textbf{Mixup}: $\tilde{x}_{p,q} \leftarrow \lambda x_p + (1 - \lambda ) x_q$ for $p,q$ pairs, $\lambda \sim Beta(0.7,0.7)$}
                \State{\textbf{Task Update}: $\theta,\sigma \leftarrow \arg\min\nolimits_{\sigma, \theta} \sum\nolimits\mathcal{L}_T(\sigma, \theta, \tilde{x}_{i,j})$ }
                \State{\textbf{Domain Update}: $\theta,\mu \leftarrow  \arg\left[\min\nolimits_{\theta} \max\nolimits_\mu - \gamma \sum\nolimits \mathcal{L}_D(\mu, \theta, \tilde{x}_{i,j})\right]$}
            \EndFor
        \EndProcedure
        \Procedure{Testing MixDANN on Target $\mathbb{Q}$}{}
                \State{\textbf{Prediction}: $y_i\leftarrow c_\sigma \circ r_\theta(x_i)$ for samples from target $x_i \sim \mathbb{Q}$}
        \EndProcedure
    \end{algorithmic}
\end{algorithm}

\vspace{5pt}
\noindent\textbf{Connecting Mixup to the $\mathcal{H}$-divergence.} From the perspective of the mentioned theory motivating DANN for DG (above), this form of density estimation has interestingly been linked to invariant learning by data-augmentation \cite{chapelle2001vicinal}. In particular, the vicinal measure $v_*$ may be defined to promote learning which is \textit{invariant} to features of our data (e.g., augmentation by noise seeks to make a neural network's prediction invariant to noise in the input). By applying Mixup as defined in Eq.~\eqref{eqn:mixup}, we coincidentally mix features across the domains $\mathbb{P}_1, \ldots, \mathbb{P}_k$ because $x_p$ and $x_q$ may be drawn from differing domains. In this sense, the proposed vicinal distribution $u_*$ estimates the density of a data point $x_i$ to promote invariance to \textit{domain} features in our learning algorithm. Hence, Mixup in the presence of multiple domains may be viewed as a technique complementary to DANN. It too is aimed at training $r_\theta$ and $c_\sigma$ to be invariant to domain features so that the errors of the domain discriminator $d_\mu$ are maximized, and subsequently, the $\mathcal{H}$-divergence is minimized. Alg.~\ref{alg:one} shows \textit{MixDANN}, our algorithmic \textit{and} theoretical combination of Mixup and DANN for DG.

% \vspace{-8pt}
\section{Experiments} % ~2 pages
\label{sec:exp}
% \vspace{-5pt}
\subsection{Experimental Setup}
\begin{figure}[t!]
    \centering
    \resizebox{0.99\columnwidth}{!}{%
    \centering 
    % \footnotesize
    \setlength{\tabcolsep}{5pt}
    \renewcommand{\arraystretch}{1.1}
    \begin{tabular}{lcccc}
    \specialrule{.1em}{.1em}{0.1em} 
    \textbf{Sites} & \textbf{$N$} & \textbf{Scanner}   & \textbf{FLAIR Voxel Size (mm$^3$)} & \textbf{TR/TE (ms)} \\ \hline
    Amsterdam      & 20           & 3T GE Signa HDxt   & $0.98 \times 0.98 \times 1.20$     & 8000/126            \\
    Singapore      & 20           & 3T Siemens TrioTim & $1.00 \times 1.00 \times 3.00$     & 9000/82             \\
    Utrecht        & 20           & 3T Philips Achieva & $0.98 \times 0.98 \times 1.20$     & 11000/125           \\ \hline
    Pittsburgh     & 20           & 3T Siemens TrioTim & $1.00 \times 1.00 \times 3.00$     & 9160/90    \\
    \specialrule{.1em}{.1em}{0.1em}         
    \end{tabular}
    }
    \begin{subfigure}[b]{0.24\columnwidth}
         \centering
         \includegraphics[height=55pt]{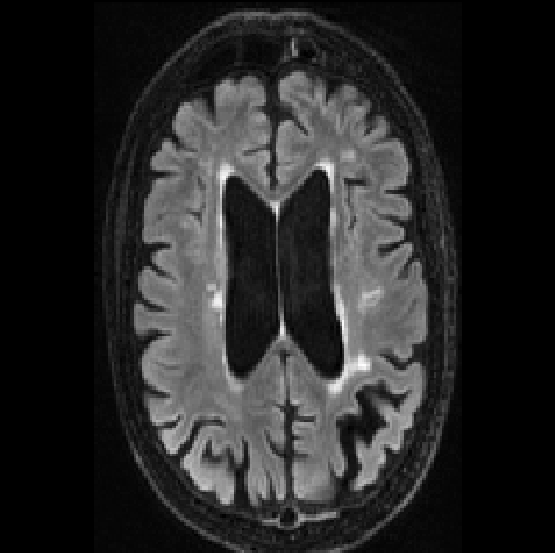}
         \vspace{-5pt}
         \caption{Amsterdam}
         \label{fig:Amsterdam}
     \end{subfigure}
    \begin{subfigure}[b]{0.24\columnwidth}
         \centering
         \includegraphics[height=55pt]{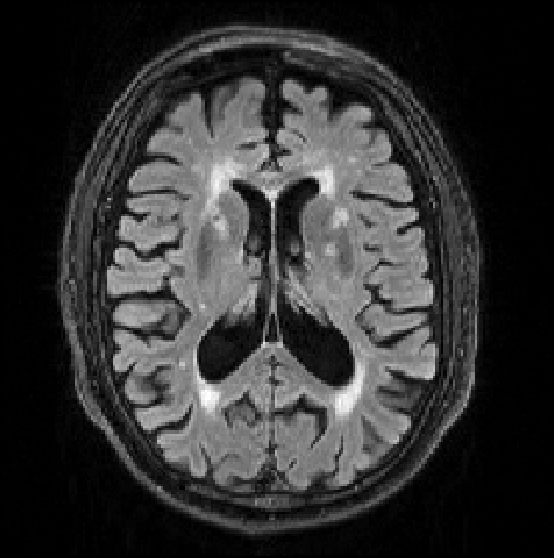}
         \vspace{-5pt}
         \caption{Singapore}
         \label{fig:Singapore}
     \end{subfigure}
    \begin{subfigure}[b]{0.24\columnwidth}
         \centering
         \includegraphics[height=55pt]{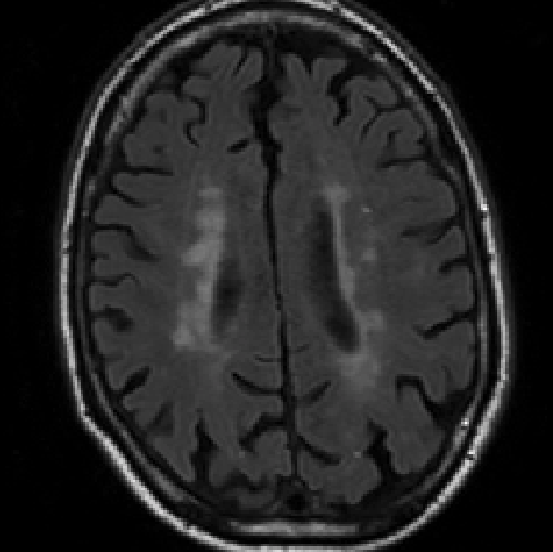}
         \vspace{-5pt}
         \caption{Utrecht}
         \label{fig:Utrecht}
     \end{subfigure}
    \begin{subfigure}[b]{0.24\columnwidth}
         \centering
         \includegraphics[height=55pt]{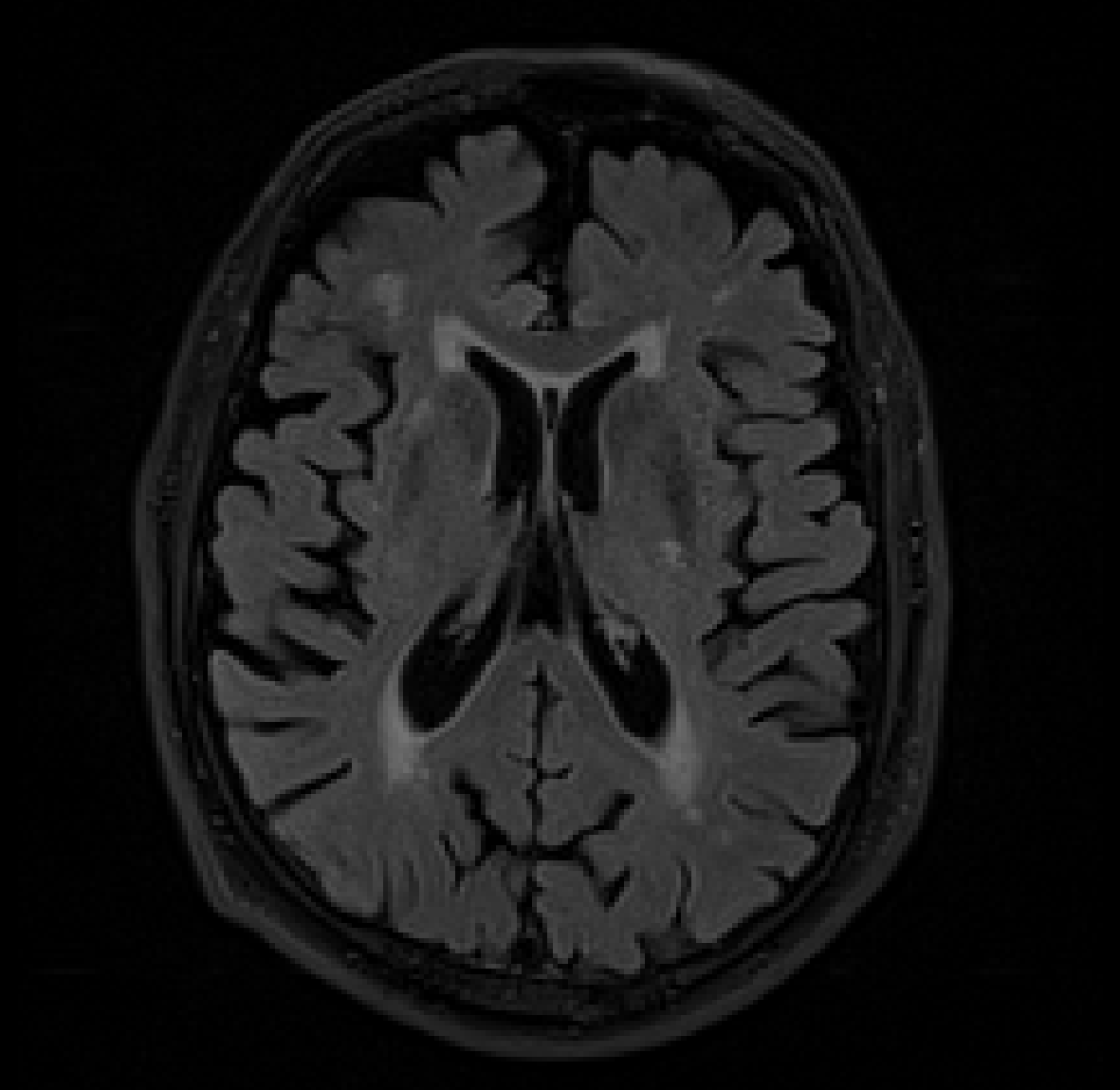}
         \vspace{-17pt}
         \caption{Pittsburgh}
         \label{fig:Pitt}
     \end{subfigure}
    %  \vspace{-8pt}
    \caption{\label{table:datasets} Top: Datasets with FLAIR and T1 (coregistered to FLAIR). Bottom: Sample FLAIR images.}
\end{figure}

% \vspace{-5pt}
% \vspace{-3pt}
\noindent\textbf{Datasets.}
We evaluate on two WMH datasets consisting of FLAIR (Fluid Attenuated Inverse Recovery), T1, and manual WMH segmentation for each subject: (1) a multi-site public MICCAI WMH Challenge Dataset \cite{kuijf2019standardized} from three sites (Amsterdam (\textbf{A}), Singapore (\textbf{S}), Utrecht (\textbf{U})) and (2) our local in-house dataset \cite{karim2019relationships} from a single site (Pittsburgh (\textbf{P})). See Fig.~\ref{table:datasets} for distinct scanners/protocols and \cite{kuijf2019standardized,karim2019relationships} for details.

\vspace{5pt}
\noindent\textbf{Metrics.}
We use five evaluation metrics to assess the WMH prediction mask $\hat{Y}$ against the ground truth segmentation mask $Y$ (TP: true positive, FP: false positive, FN: false negative). (1) Dice Similarity Coefficient (\textbf{DSC}): $2(Y \cap \hat{Y}) / (|Y| + |\hat{Y}|)$, (2) Housdorff Distance (\textbf{H95}): $H95 = \max \{ \sup\nolimits_{x \in Y} \inf\nolimits_{y \in \hat{Y}} d(x,y),\sup_{y \in Y} \inf_{x \in \hat{Y}} d(x,y)\}$ using the 95-th percentile distance, (3) Absolute Volume Difference (\textbf{AVD}) between the predicted and true WMH volume,
% $AVD = |V(Y) - V(\hat{Y})| / V(Y)$ where $V(\cdot)$ is the WMH volume, 
(4) \textbf{Lesion Recall}: Computes the \# of correctly detected WMH over the \# of true WMH, (5) \textbf{Lesion F1}: TP / (TP + 0.5(FP+FN)).
% Computes the \# of correctly detected WMH over the \# of incorrectly detected WMH.

\vspace{5pt}
\noindent\textbf{Baseline Models.} These baselines build on U-Net \cite{unet}, and we report their results by a recent DA/DG work \cite{li2020domain}: (1) \textbf{DeepAll} is the baseline U-Net with no DA or DG mechanisms which the following models build on.
(2) \textbf{UDA} \cite{li2020domain} is an unsupervised DA method using 
% DANN with 1 or 2 target scans (\textbf{Few Shot}) or 
\textit{all} target scans but not their labels.
% (\textbf{Full Set}). 
Note this \textit{DA} method \textit{requires} target thus has an advantage over DG methods. (3) \textbf{BigAug} \cite{zhang2020generalizing} is a state-of-the-art DG medical imaging segmentation method with heavy data augmentations. See \cite{li2020domain,zhang2020generalizing} for full details.

\vspace{5pt}
\noindent\textbf{Our Model.} Our proposed approaches also build on U-Net. Standard data augmentations (rotation, scale, shear) are applied. (1) We implement our own \textbf{DeepAll} comparable to the DeepAll by \cite{li2020domain} to our best effort for fair comparison. (2) \textbf{DANN}: We introduce the domain discriminator (Conv-Conv-Conv-FC-FC-FC) $d_\mu$ to U-Net to the output of the second downsampling layer (Fig.~\ref{fig:model}). For the purpose of DANN (Eq.~\eqref{eqn:dann_obj}), we can treat the U-Net as $c_\sigma \circ r_\theta$ where $r_\theta$ is before and $c_\sigma$ is after the second downsampling. We slowly introduce $\mathcal{L}_{D}$ to $r_\theta$ by setting $\gamma=(2\cdot\xi)/(1+\exp(-\kappa\cdot p))-1$ in Eq.~\eqref{eqn:dann_obj} with $p=\textrm{epoch}/\textrm{max\_epoch}$, $\kappa = 3$, and $\xi = 0.1$.  (3) \textbf{Mixup}: We randomly mix the samples following Eq.~\eqref{eqn:mixup} to induce domain invariance.
% with the mixup loss: $\mathcal{L}_T(\tilde{x}_{i,j}) = \lambda\mathcal{L}_T(\tilde{x}_{i}) + (1-\lambda)\mathcal{L}_T(\tilde{x}_{j})$.
(4) \textbf{MixDANN}: Our final model combines DANN and Mixup. The initial learning rate is 2e-4 for all models. We trained on 80\% of the training data against the comparing methods using 2 x NVIDIA RTX2080Ti. 
% The exact implementation details will be in the code released upon acceptance.

% \vspace{-5pt}
\subsection{Results and Analysis}
% \vspace{-5pt}
Each model tests on one \textit{target} domain after training on the remaining \textit{sources} (Sources$\rightarrow$Target). Generally, the comparison in DG is subjective across different experimental setups. For instance, despite the near identical architectures, our DeepAll slightly under-performs on some targets. As such, we pay special attention to the relative performance \textbf{gain} over the respective DeepAll to assess the domain generalizability.

\begin{figure}[t!]
         \centering
    \begin{subfigure}[b]{0.19\columnwidth}
         \centering
         \includegraphics[width=\textwidth]{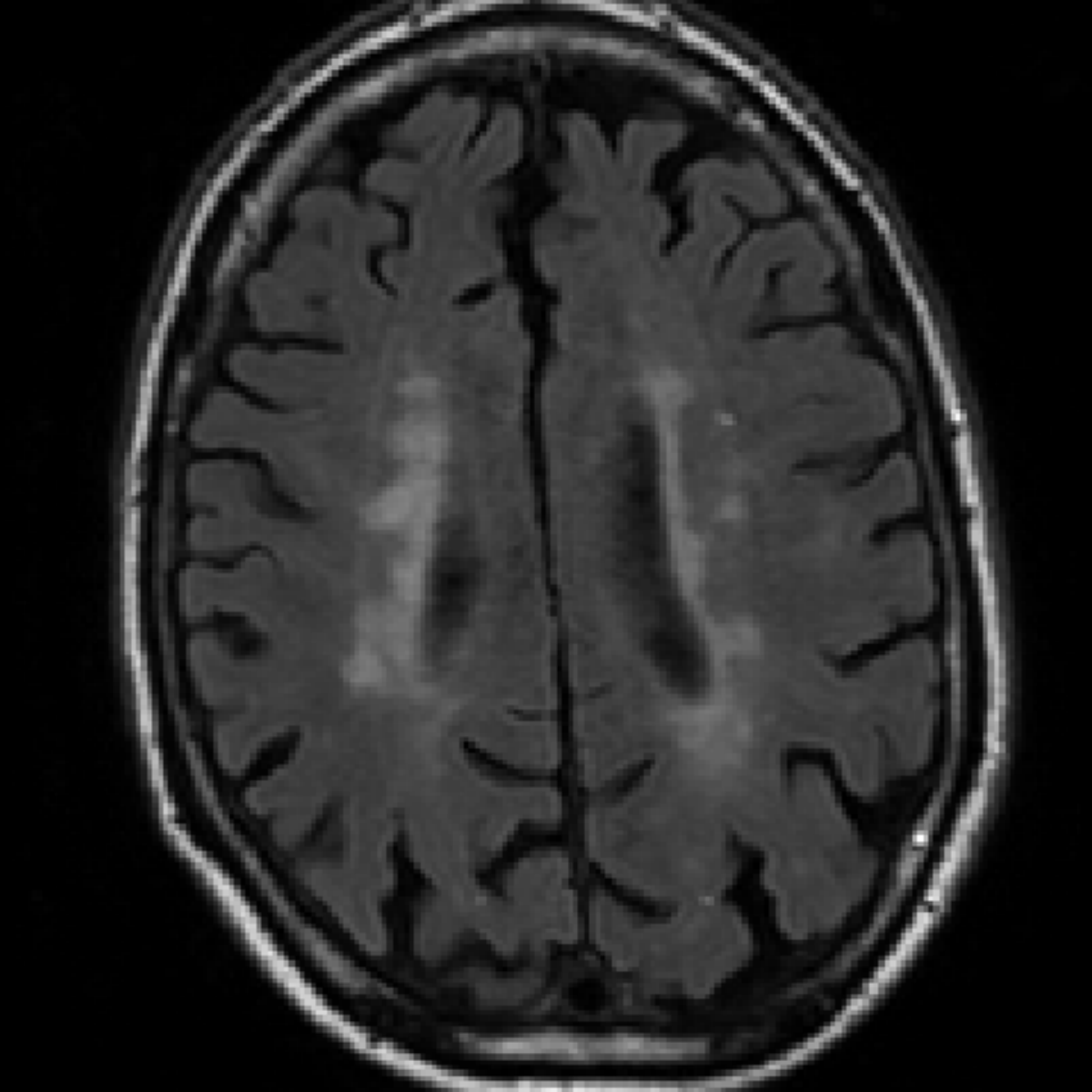}\\
         \includegraphics[width=\textwidth]{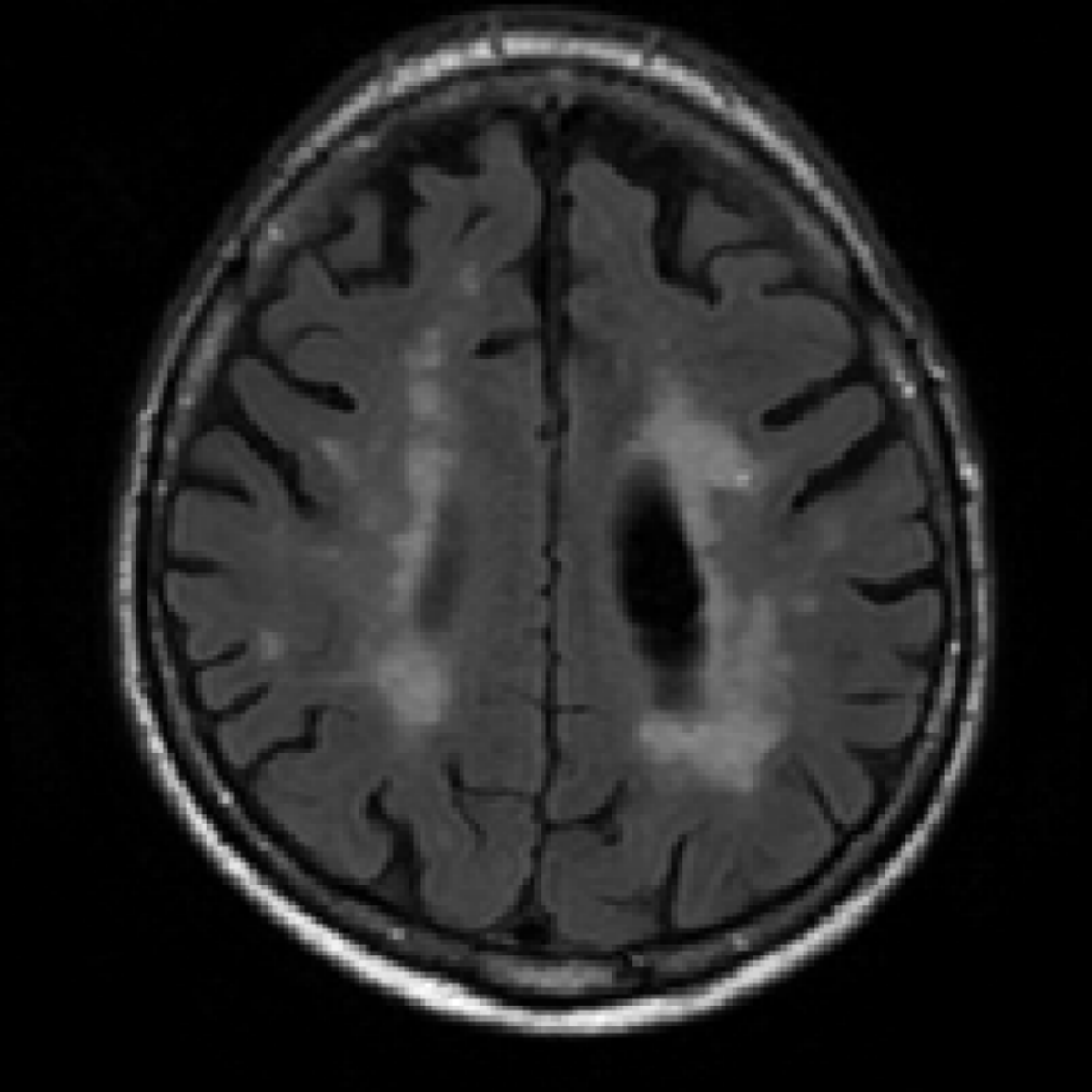}
         \vspace{-13pt}
         \caption{\scriptsize FLAIR}
     \end{subfigure}
         \centering
    \begin{subfigure}[b]{0.19\columnwidth}
         \centering
         \includegraphics[width=\textwidth]{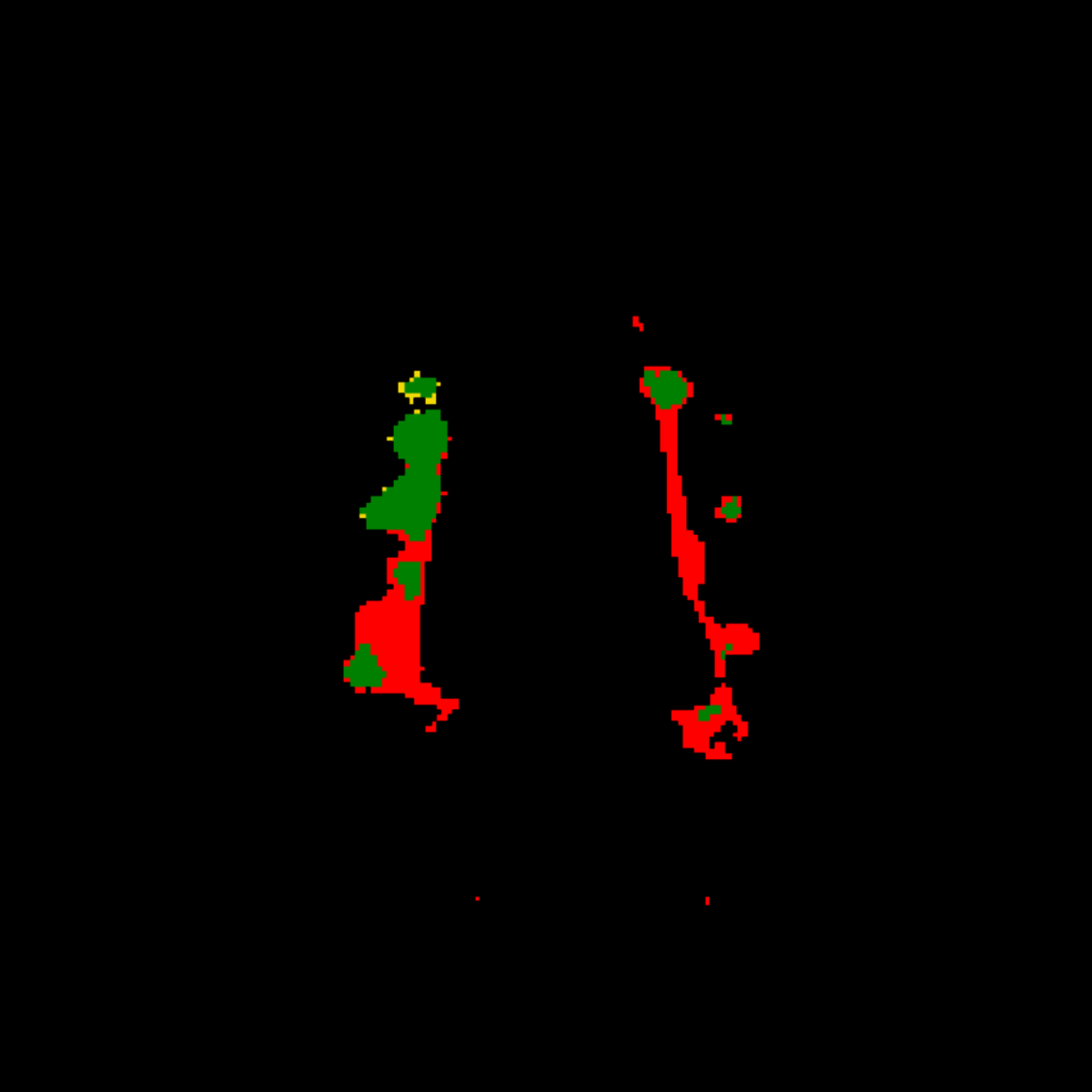}\\
         \includegraphics[width=\textwidth]{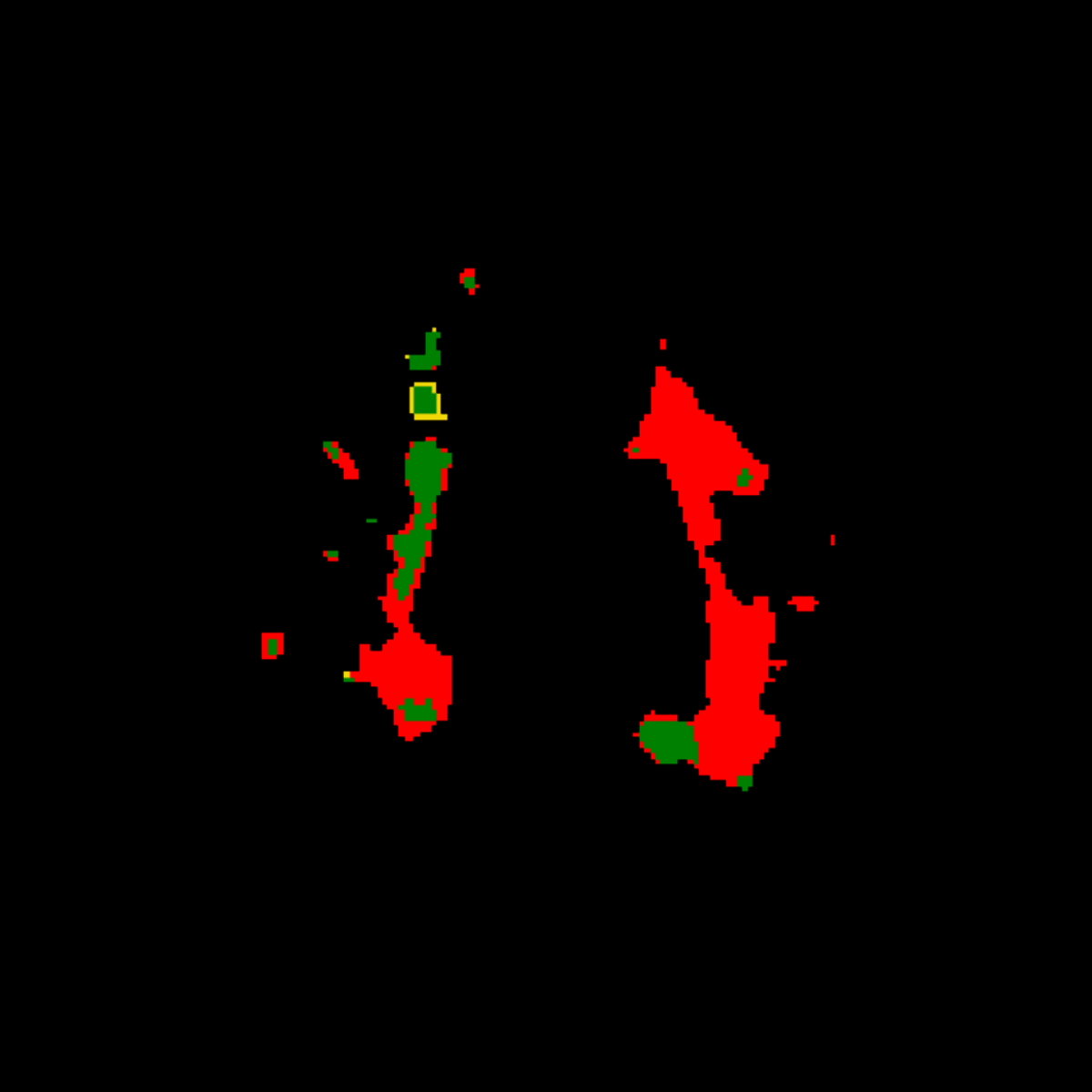}
         \vspace{-13pt}
         \caption{\scriptsize DeepAll}
     \end{subfigure}
    \begin{subfigure}[b]{0.19\columnwidth}
         \centering
         \includegraphics[width=\textwidth]{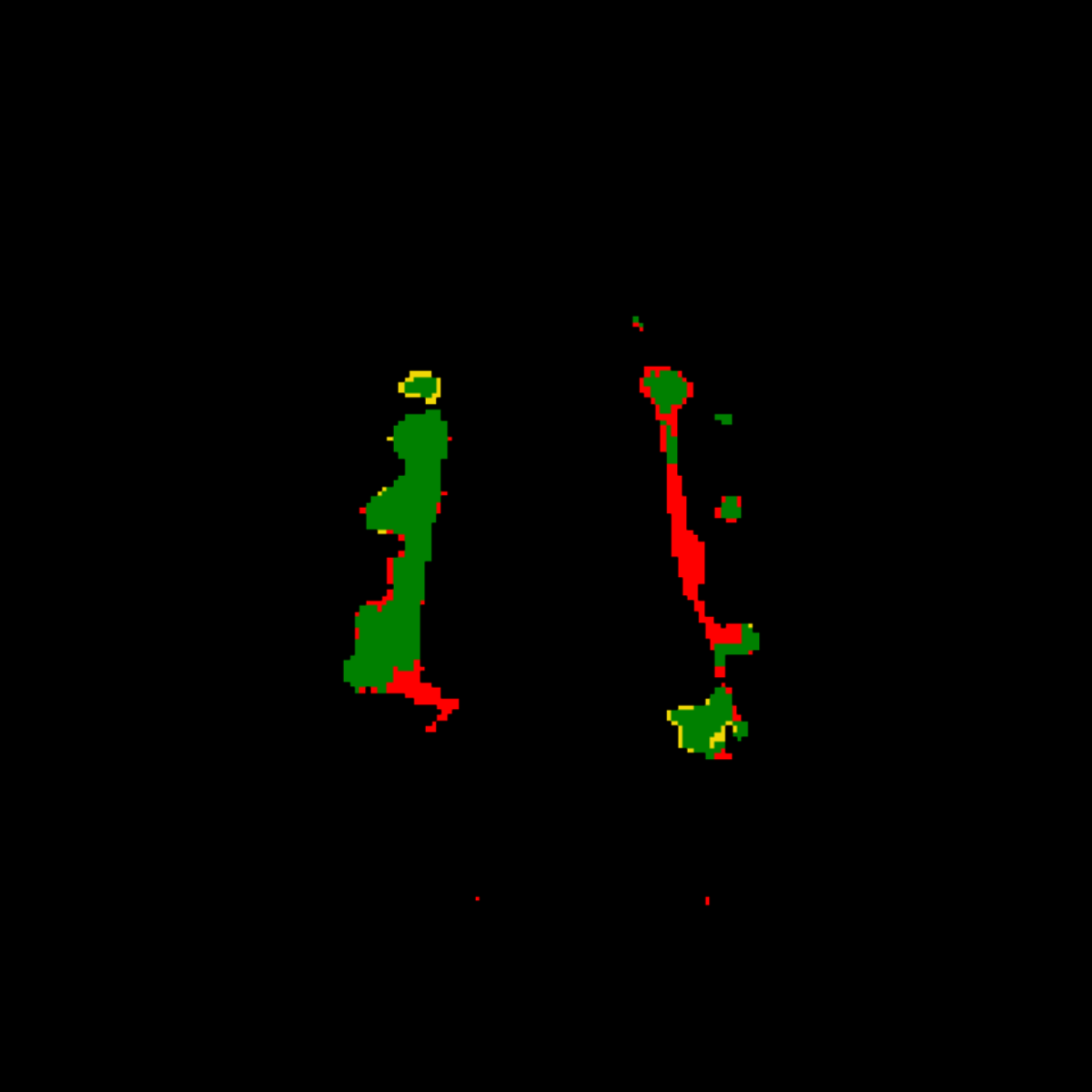}
         \includegraphics[width=\textwidth]{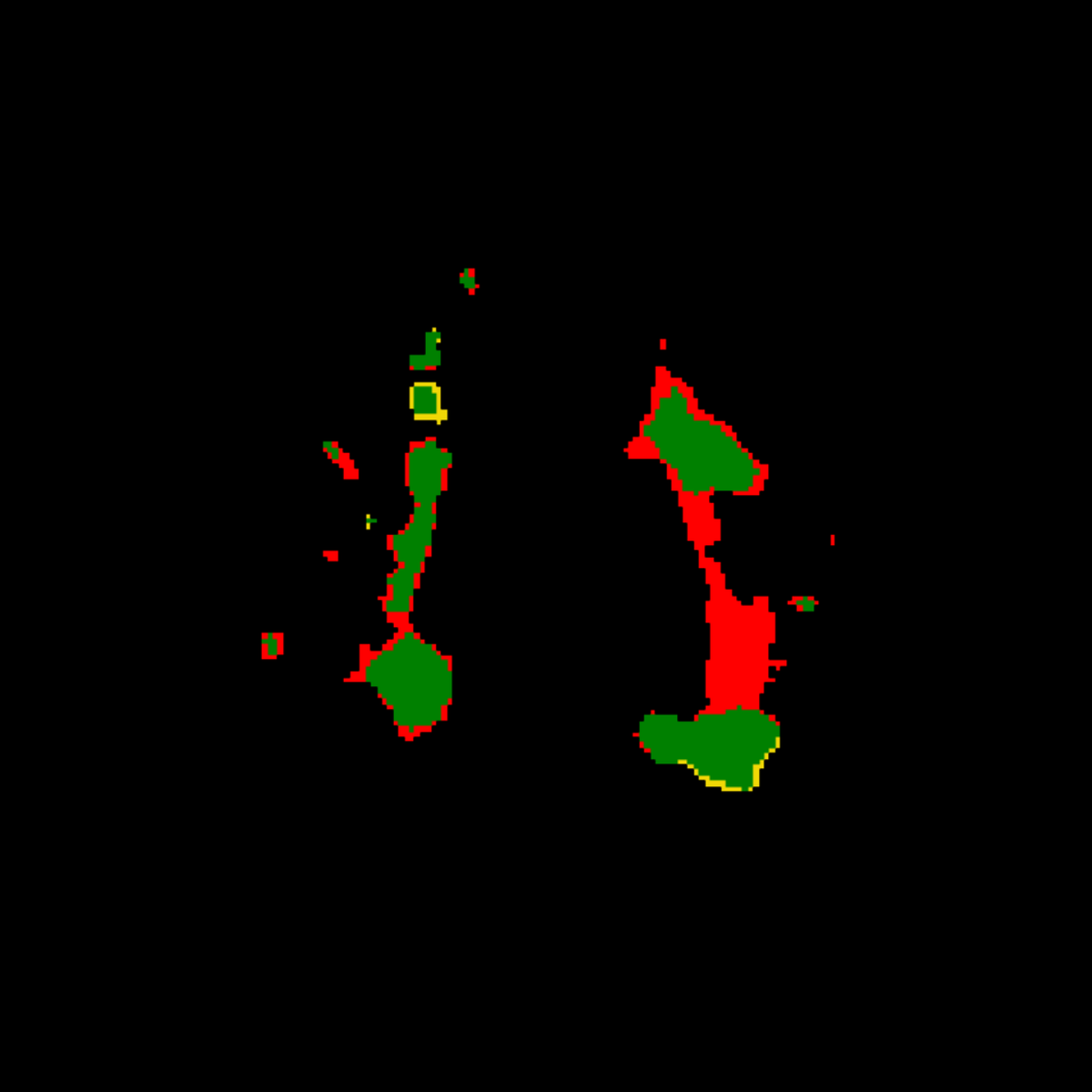}
         \vspace{-13pt}
         \caption{\scriptsize DANN}
     \end{subfigure}
    \begin{subfigure}[b]{0.19\columnwidth}
         \centering
         \includegraphics[width=\textwidth]{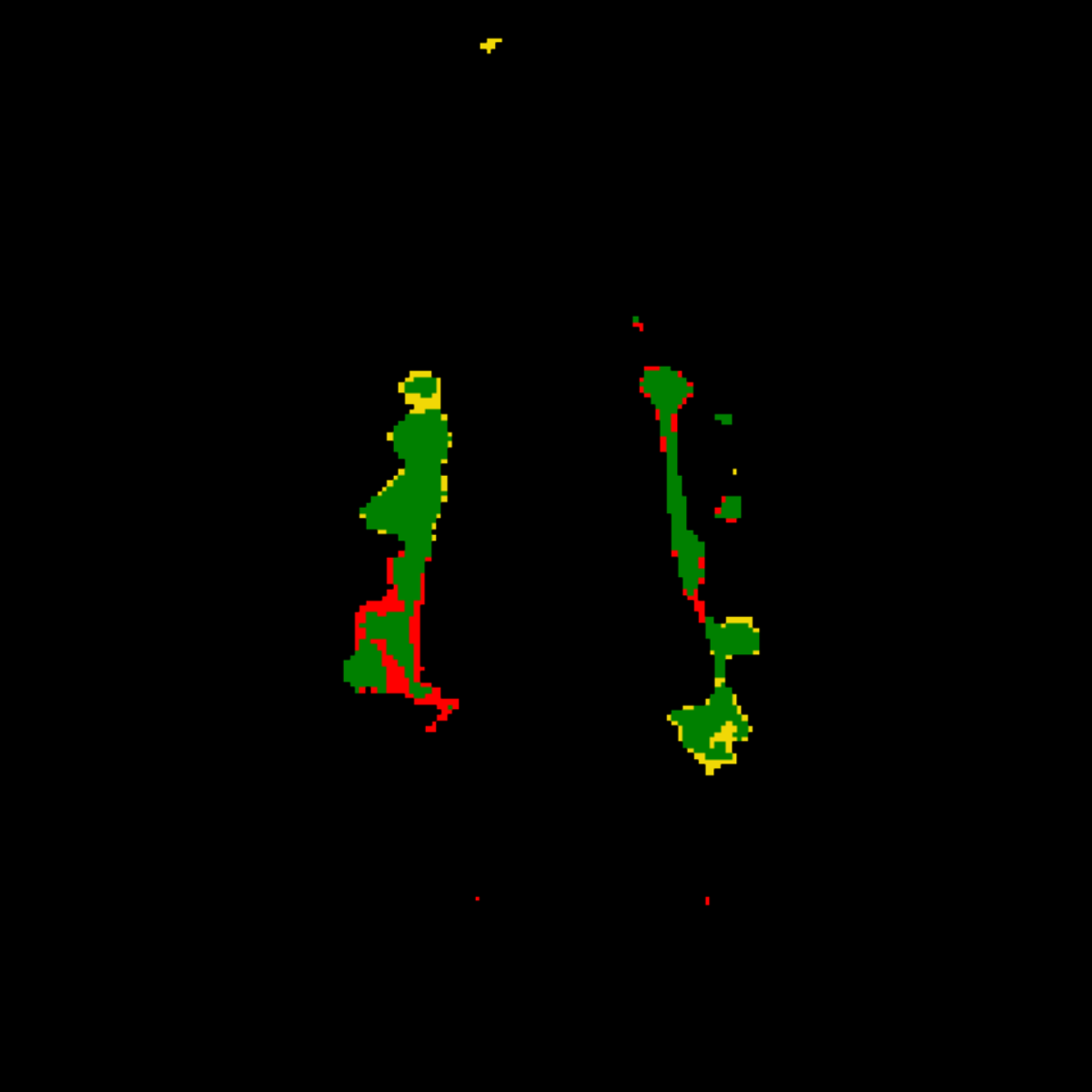}\\
         \includegraphics[width=\textwidth]{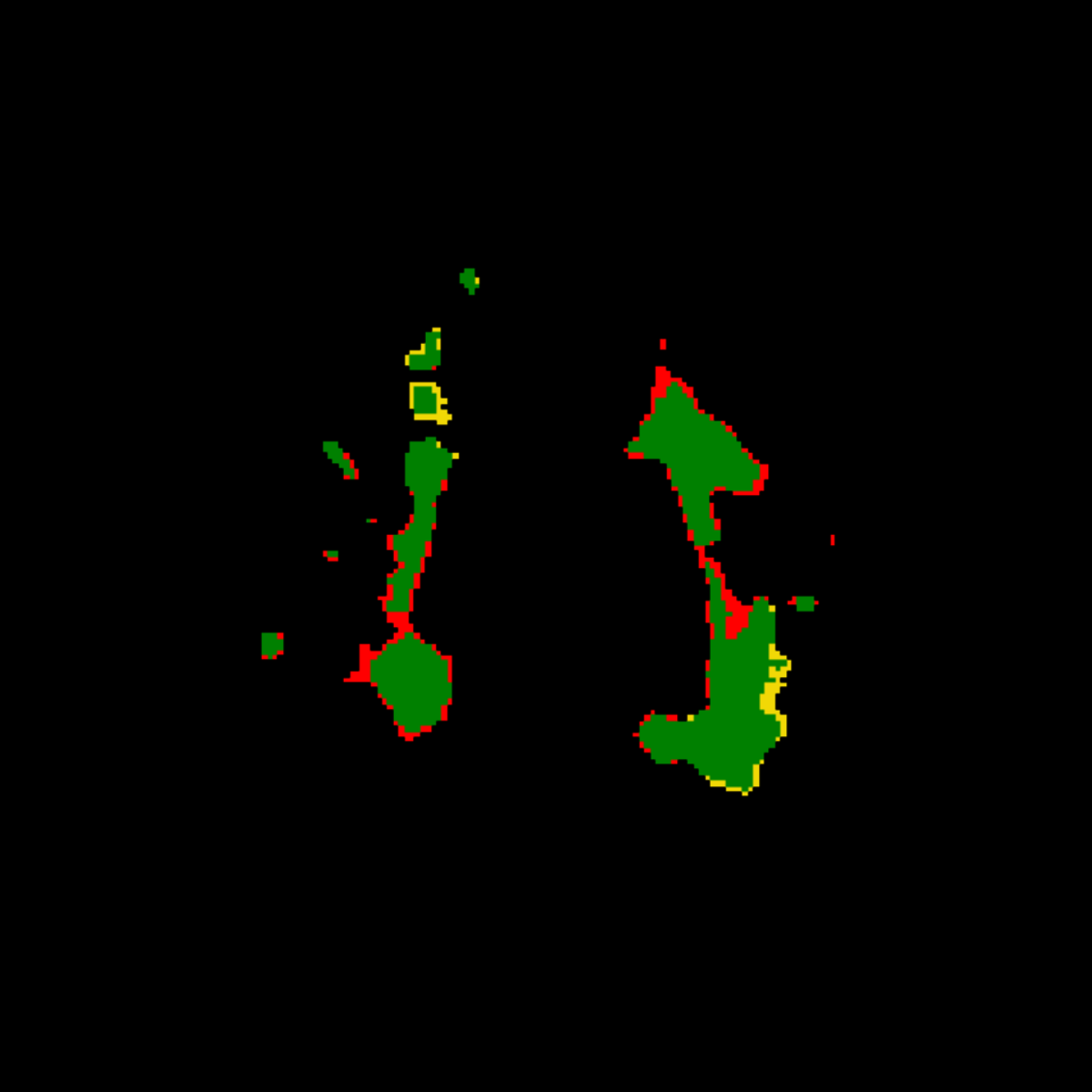}
         \vspace{-13pt}
         \caption{\scriptsize Mixup}
     \end{subfigure}
    \begin{subfigure}[b]{0.19\columnwidth}
         \centering
         \includegraphics[width=\textwidth]{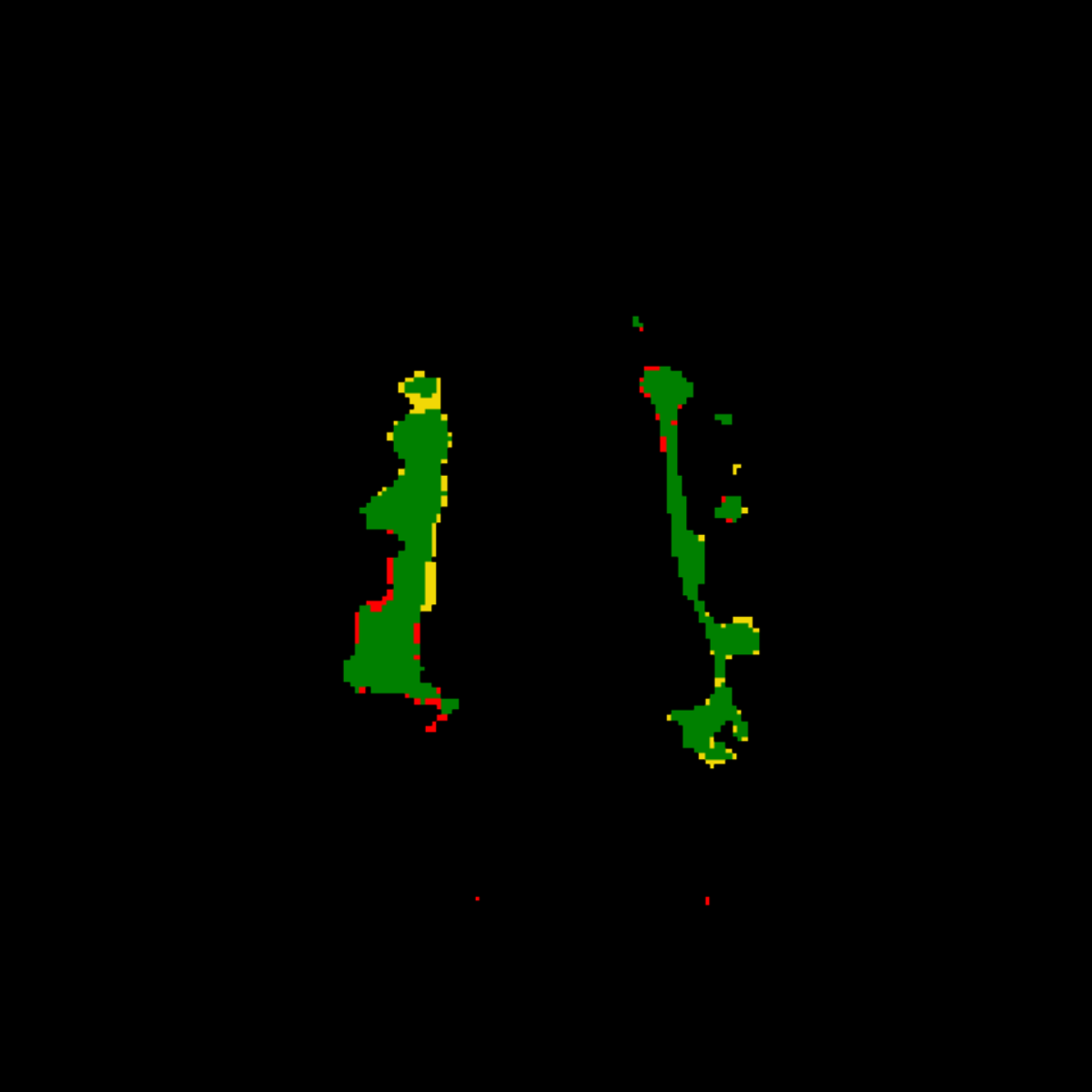}\\
         \includegraphics[width=\textwidth]{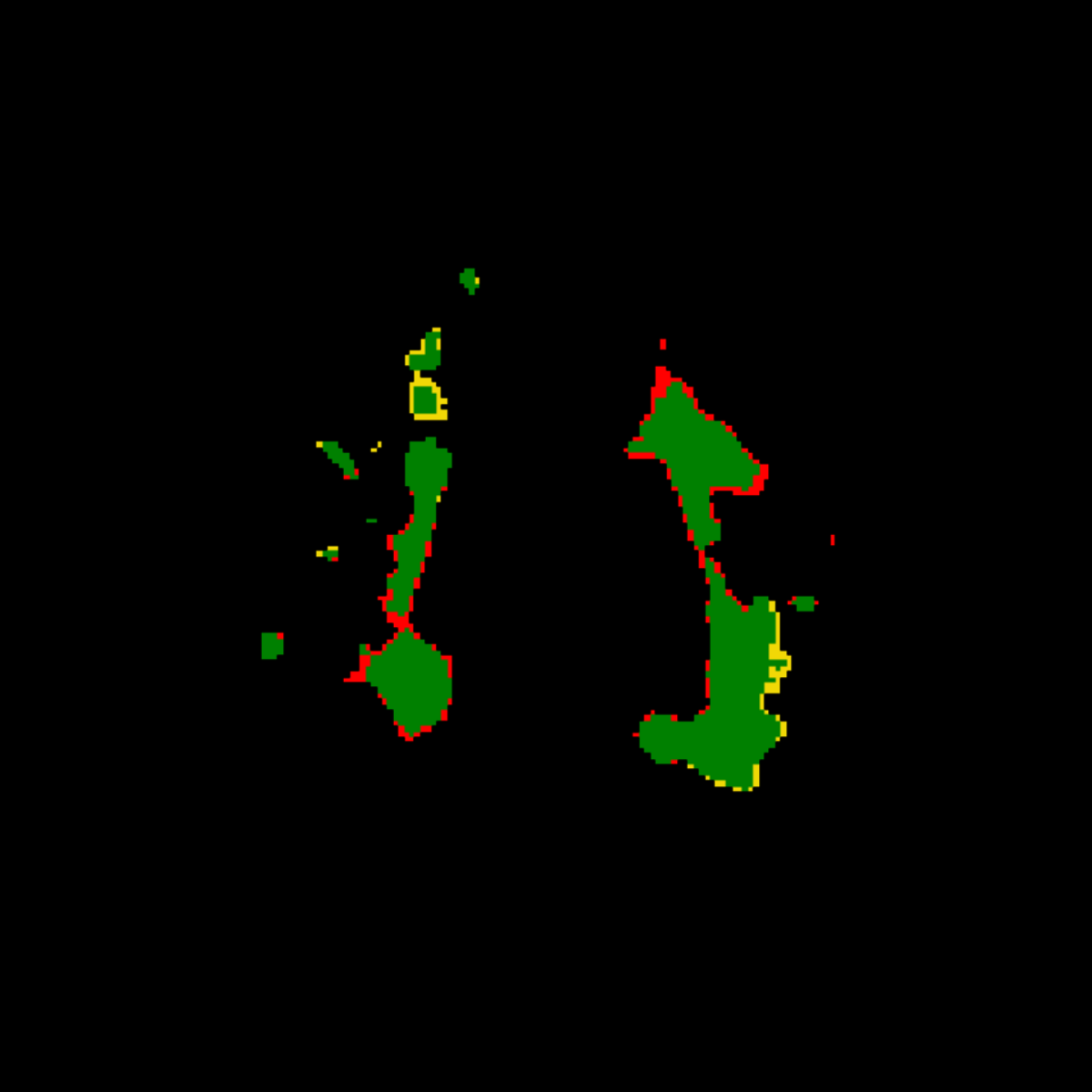}
         \vspace{-13pt}
         \caption{\scriptsize MixDANN}
     \end{subfigure}
    %  \vspace{-8pt}
    \caption{\label{fig:wmh_visual} Visualization of WMH predictions (\textbf{A+S$\rightarrow$U}). True Positive: Green, False Positive: Yellow, False Negative: Red.}
\end{figure}

\vspace{5pt}
\noindent\textbf{Exp 1: DG within WMH Challenge Dataset.} We test on each target by training on the remaining two sources.  Table~\ref{table:results_whm} shows the results of all targets and the average across them. We see that despite our weaker DeepAll, MixDANN shows the best absolute \textbf{avg} on three metrics (DSC, AVD, and Lesion F1) and the best relative \textbf{gain} on all metrics. We ablate and see improvements in the order of DANN, Mixup, and MixDANN, also visualized in Fig.~\ref{fig:wmh_visual}. We note that \textbf{AVD} assessing the accuracy of the predicted WMH volume, which is often considered as a biomarker of vascular pathology, is most accurate by MixDANN. We also pay special attention to the improvement in the hardest case of \textbf{A+S}$\rightarrow$\textbf{U}: This \textit{exactly} exemplifies a possible scenario where a DeepAll may fail on an unseen dataset but MixDANN can assure robustness.

\begin{table}[t!]
\caption{\label{table:results_whm} WMH Challenge Results. \textbf{Sources}$\rightarrow$\textbf{Target}: Trained on \textbf{Sources} and tested on \textbf{Target} among  \textbf{A}msterdam, \textbf{S}ingapore, and \textbf{U}trecht. \textbf{avg}: Average of the target results. \textbf{gain}: \textbf{avg} gain over the respective DeepAll.}
% \vspace{-7pt}
\resizebox{\columnwidth}{!}{%
\centering 
% \footnotesize
\setlength{\tabcolsep}{4pt}
\renewcommand{\arraystretch}{1.1}
\begin{tabular}{lccccc}
\specialrule{.1em}{.1em}{0.1em} 
                \textbf{Model}    & \textbf{A+S$\rightarrow$U} & \textbf{U+S$\rightarrow$A} & \multicolumn{1}{c|}{\textbf{A+U$\rightarrow$S}} & \textbf{avg}   & \textbf{gain}    \\ \hline
\multicolumn{6}{c}{\textbf{DSC $\uparrow$} (Higher is better)}                                                                                                        \\ \hline
DeepAll (\cite{li2020domain} Setup)          & 0.430            & 0.674              & \multicolumn{1}{c|}{0.682}              & 0.595          & -                \\
BigAug \cite{zhang2020generalizing}            & 0.534            & 0.691              & \multicolumn{1}{c|}{0.711}              & 0.645          & 0.050            \\
% UDA (Few Shot)    & 0.489            & 0.733              & \multicolumn{1}{c|}{0.780}              & 0.667          & 0.072            \\
UDA \cite{li2020domain} (Full Set)  & 0.529            & \textbf{0.737}     & \multicolumn{1}{c|}{0.782}              & 0.683          & 0.087            \\ \hline
DeepAll (Our Setup)           & 0.183            & 0.619              & \multicolumn{1}{c|}{0.781}              & 0.528          & -                \\
DANN              & 0.315            & 0.674              & \multicolumn{1}{c|}{0.773}              & 0.587          & 0.060            \\
Mixup             & 0.619            & 0.691              & \multicolumn{1}{c|}{0.835}              & 0.715          & 0.187            \\
MixDANN      & \textbf{0.694}   & 0.700              & \multicolumn{1}{c|}{\textbf{0.839}}     & \textbf{0.744} & \textbf{0.217}   \\ \hline
\multicolumn{6}{c}{\textbf{H95 $\downarrow$} (Lower is better)}                                                                                      \\ \hline
DeepAll          & 11.46            & 11.51              & \multicolumn{1}{c|}{9.22}               & 10.73          & -                \\
BigAug            & \textbf{9.49}    & 9.77               & \multicolumn{1}{c|}{8.25}               & 9.17           & -1.56            \\
% UDA (Few Shot)    & 11.02            & 7.90               & \multicolumn{1}{c|}{7.54}               & 8.82           & -1.91            \\
UDA (Full Set) & 10.01            & \textbf{7.53}      & \multicolumn{1}{c|}{7.51}               & \textbf{8.35}  & -2.38            \\ \hline
DeepAll          & 42.69            & 18.05              & \multicolumn{1}{c|}{4.56}               & 21.77          & -                \\
DANN              & 38.56            & 15.48              & \multicolumn{1}{c|}{5.15}               & 19.73          & -2.03            \\
Mixup             & 24.08            & 13.21              & \multicolumn{1}{c|}{5.70}               & 14.33          & -7.44            \\
MixDANN      & 20.57            & 12.75              & \multicolumn{1}{c|}{\textbf{3.10}}      & 12.14          & \textbf{-9.63}   \\ \hline
\multicolumn{6}{c}{\textbf{AVD $\downarrow$} (Lower is better)}                                                                                      \\ \hline
DeepAll           & 54.84            & 37.60              & \multicolumn{1}{c|}{45.95}              & 46.13          & -                \\
BigAug            & 47.46            & 30.64              & \multicolumn{1}{c|}{35.41}              & 37.84          & -8.29            \\
% UDA (Few Shot)    & 57.01            & \textbf{16.01}     & \multicolumn{1}{c|}{24.75}              & 32.59          & -13.54           \\
UDA (Full Set) & 54.95            & 30.97              & \multicolumn{1}{c|}{22.14}              & 36.02          & -10.11           \\ \hline
DeepAll           & 384.19           & 43.28              & \multicolumn{1}{c|}{23.26}              & 150.24         & -                \\
DANN              & 134.03           & 26.65              & \multicolumn{1}{c|}{24.09}              & 61.59          & -88.66           \\
Mixup             & 42.09            & 33.47              & \multicolumn{1}{c|}{13.41}              & 29.66          & -120.59          \\
MixDANN      & \textbf{23.40}   & \textbf{26.48}              & \multicolumn{1}{c|}{\textbf{12.81}}     & \textbf{20.89} & \textbf{-129.35} \\ \hline
\multicolumn{6}{c}{\textbf{Lesion Recall $\uparrow$} (Higher is better)}                                                                                              \\ \hline
DeepAll           & 0.634            & 0.692              & \multicolumn{1}{c|}{0.641}              & 0.656          & -                \\
BigAug            & 0.643            & 0.709              & \multicolumn{1}{c|}{0.691}              & 0.681          & 0.025            \\
% UDA (Few Shot)    & 0.639            & 0.785              & \multicolumn{1}{c|}{0.666}              & 0.697          & 0.041            \\
UDA (Full Set) & \textbf{0.652}   & \textbf{0.841}     & \multicolumn{1}{c|}{0.754}              & \textbf{0.749} & 0.093            \\ \hline
DeepAll           & 0.309            & 0.623              & \multicolumn{1}{c|}{0.705}              & 0.546          & -                \\
DANN              & 0.349            & 0.630              & \multicolumn{1}{c|}{0.740}              & 0.573          & 0.028            \\
Mixup             & 0.556            & 0.700              & \multicolumn{1}{c|}{0.790}              & 0.682          & 0.136            \\
MixDANN      & 0.604            & 0.685              & \multicolumn{1}{c|}{\textbf{0.797}}     & 0.695          & \textbf{0.150}   \\ \hline
\multicolumn{6}{c}{\textbf{Lesion F1 $\uparrow$} (Higher is better)}                                                                                                  \\ \hline
DeepAll           & 0.561            & 0.673              & \multicolumn{1}{c|}{0.592}              & 0.609          & -                \\
BigAug            & 0.577            & 0.704              & \multicolumn{1}{c|}{0.651}              & 0.644          & 0.035            \\
% UDA (Few Shot)    & 0.533            & 0.725              & \multicolumn{1}{c|}{0.657}              & 0.638          & 0.030            \\
UDA (Full Set) & 0.546            & \textbf{0.739}     & \multicolumn{1}{c|}{0.649}              & 0.645          & 0.036            \\ \hline
DeepAll           & 0.288            & 0.554              & \multicolumn{1}{c|}{0.697}              & 0.513          & -                \\
DANN              & 0.309            & 0.610              & \multicolumn{1}{c|}{0.708}              & 0.542          & 0.029            \\
Mixup             & 0.515            & 0.642              & \multicolumn{1}{c|}{0.724}              & 0.627          & 0.114            \\
MixDANN      & \textbf{0.602}   & 0.651              & \multicolumn{1}{c|}{\textbf{0.728}}     & \textbf{0.660} & \textbf{0.147}  \\
\specialrule{.1em}{.1em}{0.1em} 
\end{tabular}
}
\end{table}

\begin{table}[t!]
    \vspace{-5pt}
\caption{\label{table:result_pitt} WMH Segmentation Results of \textbf{A+S+U}$\rightarrow$\textbf{Pitt}}
\vspace{-8pt}
\resizebox{\columnwidth}{!}{%
\centering 
% \footnotesize
\setlength{\tabcolsep}{8pt}
\renewcommand{\arraystretch}{1.1}
\begin{tabular}{lccccc}
\specialrule{.1em}{.1em}{0.1em} 
\textbf{Model} & \textbf{DSC $\uparrow$}   & \textbf{H95 $\downarrow$}   & \textbf{AVD $\downarrow$}   & \textbf{Recall $\uparrow$} & \textbf{F1 $\uparrow$}    \\ \hline
DeepAll        & 0.434          & 18.49          & 68.56          & 0.543           & 0.630          \\
DANN           & \textbf{0.499} & 16.04          & \textbf{62.31} & \textbf{0.622}  & \textbf{0.680} \\
Mixup          & 0.462          & 16.92          & 65.92          & 0.501           & 0.606          \\
MixDANN   & 0.488          & \textbf{15.93} & 63.41          & 0.466           & 0.566     \\
\specialrule{.1em}{.1em}{0.1em}       
\end{tabular}
    % \vspace{-15pt}
}
\end{table}

\vspace{5pt}
\noindent\textbf{Exp 2: A+S+U$\rightarrow$Pitt.} We do a ``cross-dataset'' DG: train on A+S+U and test on Pitt. We could not include Pitt as a source since it only has 5 consecutive slices of manual segmentation available. Nonetheless, when we test it as a target as shown in Table~\ref{table:result_pitt}, we again see consistent improvements over DeepAll. Interestingly, DANN best performs, implying that Mixup and DANN may also individually bring benefits. 
Our MixDANN with a \textit{single} U-Net is now ranked 6$^{th}$ on the leader board of the WMH Challenge \cite{kuijf2019standardized}, competitive against other top \textit{ensemble} U-Nets.
We observe poor performance by Mixup on the WMH counting metrics (Recall and F1), suspecting this to be from the different manual annotation standards between the two datasets. We consider this as our future work.

% \vspace{-13pt}
% \noindent\textbf{Ablation Study.} We see improvements in the order of DANN, Mixup, and MixDANN in Table~\ref{table:results_whm} and Fig.~\ref{fig:wmh_visual}. 

\begin{figure}[h!]
    \centering
    % \vspace{-5pt}
    \includegraphics[width=1\columnwidth]{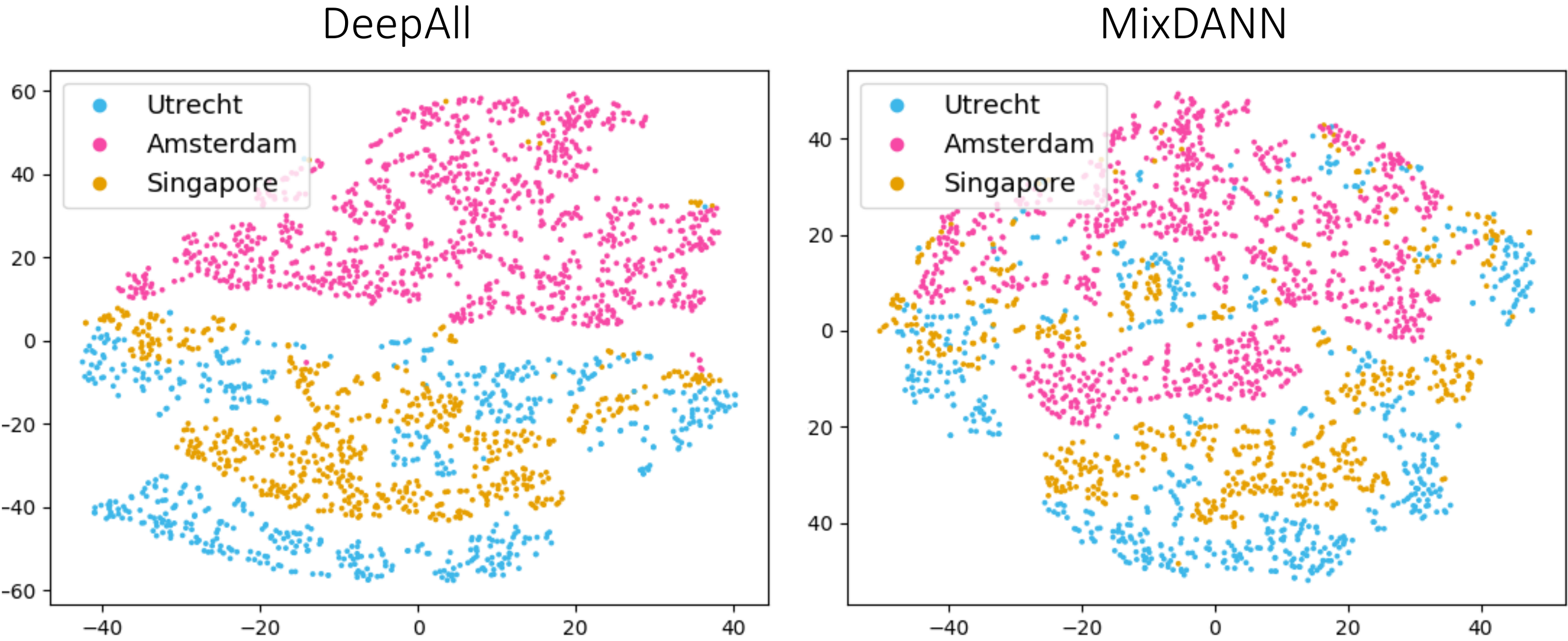}
    % \begin{subfigure}[b]{0.48\columnwidth}
    %      \centering
    %      \includegraphics[trim={1cm 0.8cm 1cm 1cm},clip,width=\textwidth]{figs/tsne/deepall_tsne_plot_perp50.png}
    % \vspace{-15pt}
    %      \caption{DeepAll}
    %      \label{fig:tsne_deepall}
    %  \end{subfigure}
    % % \begin{subfigure}[b]{0.48\columnwidth}
    % %      \centering
    % %      \includegraphics[trim={1cm 1cm 1cm 1cm},clip,width=\textwidth]{figs/tsne/dann_tsne_plot_perp50.png}
    % %      \caption{DANN}
    % %      \label{fig:tsne_dann}
    % %  \end{subfigure}
    % % \begin{subfigure}[b]{0.48\columnwidth}
    % %      \centering
    % %      \includegraphics[trim={1cm 1cm 1cm 1cm},clip,width=\textwidth]{figs/tsne/mixup_tsne_plot_perp50.png}
    % %      \caption{Mixup}
    % %      \label{fig:tsne_mixup}
    % %  \end{subfigure}
    % \begin{subfigure}[b]{0.48\columnwidth}
    %      \centering
    %      \includegraphics[trim={1cm 0.8cm 1cm 1cm},clip,width=\textwidth]{figs/tsne/dann_mixup_tsne_plot_perp50.png}
    % \vspace{-15pt}
    %      \caption{MixDANN}
    %      \label{fig:tsne_mixup_dann}
    %  \end{subfigure}
    %  \vspace{-8pt}
     \caption{\label{fig:tsne} t-SNE plots of U-Net features trained on Amsterdam (pink), Singapore (gold) and Utrecht (blue) as sources.}
    %  \vspace{-15pt}
    \vspace{5pt}
\end{figure}

\vspace{5pt}
\noindent\textbf{Do we learn domain-invariance?}
Fig.~\ref{fig:tsne} shows the t-SNE \cite{tsne} plots of the second downsampling layer ($r_\theta$) output. We see that the features by DeepAll can easily identify certain domains while those by MixDANN blur the boundary among domains as intended. In the context of WMH prediction, MixDANN explicitly suppresses the site/scanner dependent information, thus is more robust when test on unseen data.

% \vspace{-8pt}
\section{Conclusion} %
\label{sec:conclusion}
\vspace{-5pt}
We investigate the domain generalizability of a WMH segmentation deep model to be trained on sources and operate well on an unseen target.
We identify a theoretical connection between two DG approaches, namely DANN and Mixup, and jointly incorporate them into U-Net.
Using a multi-site WMH dataset and our local dataset, we show our domain invariant learning frameworks bring drastic improvements over other DA/DG methods in both relative and absolute performances.

% Can be on the 5th page from here
% \clearpage
% \section{Acknowledgements}
\label{sec:ack}
% \clearpage
% \noindent\textbf{Acknowledgments.} 
\section{Acknowledgments}
% \vspace{-12pt}
This work was supported by the NIH/NIA 
(R01 AG063752, RF1 AG025516, P01 AG025204, K23 MH118070), and SCI UR 
Scholars Award. We report no conflicts of interests.
\vspace{-0pt}

% \vspace{-12pt}
\section{Compliance with Ethical Standards}
% \vspace{-12pt}
\label{sec:ethics}
The WMH Challenge study was conducted retrospectively using open access data by WMH Segmentation Challenge (https://wmh.isi.uu.nl/). Ethical approval was not required as confirmed by its license. The Pitt study performed in line with the principles of the Declaration of Helsinki was approved by the Ethics Committee of the University of Pittsburgh.

% \vspace{-10pt}

% References should be produced using the bibtex program from suitable
% BiBTeX files (here: strings, refs, manuals). The IEEEbib.bst bibliography
% style file from IEEE produces unsorted bibliography list.
% -------------------------------------------------------------------------
\bibliographystyle{IEEEbib}
\bibliography{main}

\end{document}